\documentclass{article}
\usepackage{spconf,amsmath,graphicx,multirow,booktabs,tabularx}
\usepackage{makecell}
\usepackage{graphicx}
\usepackage{textcomp}
\usepackage{array} 
\usepackage{longtable}
\usepackage{booktabs}
\usepackage{float}
\usepackage{anyfontsize}

\usepackage{algorithm}  
\usepackage{algorithmicx}  
\usepackage{algpseudocode}  
\usepackage{amsmath} 
\usepackage{amsfonts}
\usepackage{url}

\title{Revisit out-of-vocabulary problem for slot filling : a Unified contrastive framework with multi-level data augmentations}

\name{Daichi Guo$^{1*}$, Guanting Dong$^{1*}$, Dayuan Fu$^1$, Yuxiang Wu$^1$, Chen Zeng$^1$, Tingfeng Hui$^1$ , \\
 {\em Liwen Wang$^1$, 
 Xuefeng Li$^1$,  Zechen Wang$^1$,   Keqing He$^2$, Xinyue Cui$^1$, Weiran Xu$^{1*}$}
 \thanks{$^*$The first two authors contribute equally. Weiran Xu is the corresponding author.}}
\address{$^1$Beijing University of Posts and Telecommunications,  China\\
        $^2$Meituan Group, Beijing, China}

\begin{document}
%
\maketitle
\begin{abstract}
In real dialogue scenarios, the existing slot filling model, which tends to memorize entity patterns, has a significantly reduced generalization facing Out-of-Vocabulary (OOV) problems. To address this issue, we propose an OOV robust slot filling model based on multi-level data augmentations to solve the OOV problem from both word and slot perspectives. We present a unified contrastive learning framework, which pull representations of the origin sample and augmentation samples together, to make the model resistant to OOV problems. We evaluate the performance of the model from some specific slots and carefully design test data with OOV word perturbation to further demonstrate the effectiveness of OOV words. Experiments on two datasets show that our approach outperforms the previous sota methods in terms of both OOV slots and words.

\end{abstract}
\section{Introduction}
\label{sec:intro}

Slot Filling task aims to identify semantic constituent parts from user utterances in a task-oriented dialogue system, which is a critical component of spoken language understanding(SLU). 

\begin{figure}[t]
\centering

\resizebox{.37\textwidth}{!}{\includegraphics{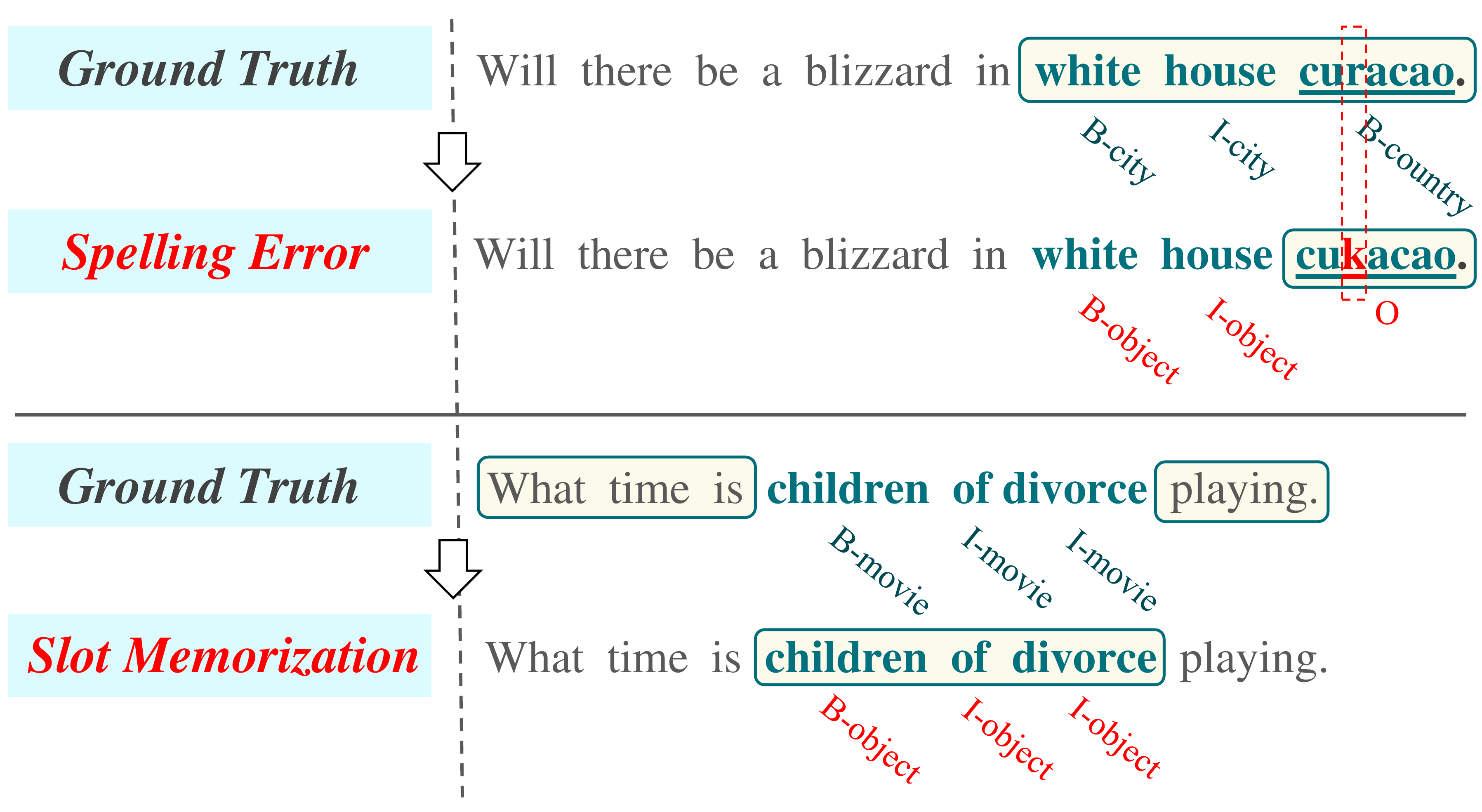}}
\caption{The illustration of the baseline model suffering from OOV word problem and OOV slot problem in slot filling task.}
\label{fig:intro}
\end{figure}

Several neural-based slot filling models\cite{liu2016attention}\cite{goo2018slot}\cite{he2020multi}\cite{wang2022instructionner}have achieved great performance due to continuous development of deep learning. While these models work well on traditional datasets, the distribution of data in the real circumstances is more variable and complex. On the one hand, there are kinds of \textbf{Out-of-Vocabulary words} in real-world corpora, such as rare words, typos words and domain-specific words. Model performance will be extremely degraded facing OOV words since they are usually not seen in the training stage. As shown in Figure \ref{fig:intro}, "curacao" becomes "cukacao" after misspelled a single letter and the previous slot filling models fail to recognize its label “B-country”. Even the predictions for the previous slot could be affected. 
On the other hand, the values of some slots, such as movie name and music name, have no restrictions on length and language patterns. Besides, the usage of words in these slots is totally different from the meaning inherent in themselves. We name this kind of slot as \textbf{Out-of-Vocabulary slot}. 
As shown in Figure \ref{fig:intro}, the previous slot filling model mistakenly recognize "children of divorce" as "object\_name" slot type because words in this phrase are labeled as "object\_name" in the training dataset.
 
\begin{figure}[t]
\centering

\resizebox{.35\textwidth}{!}{\includegraphics{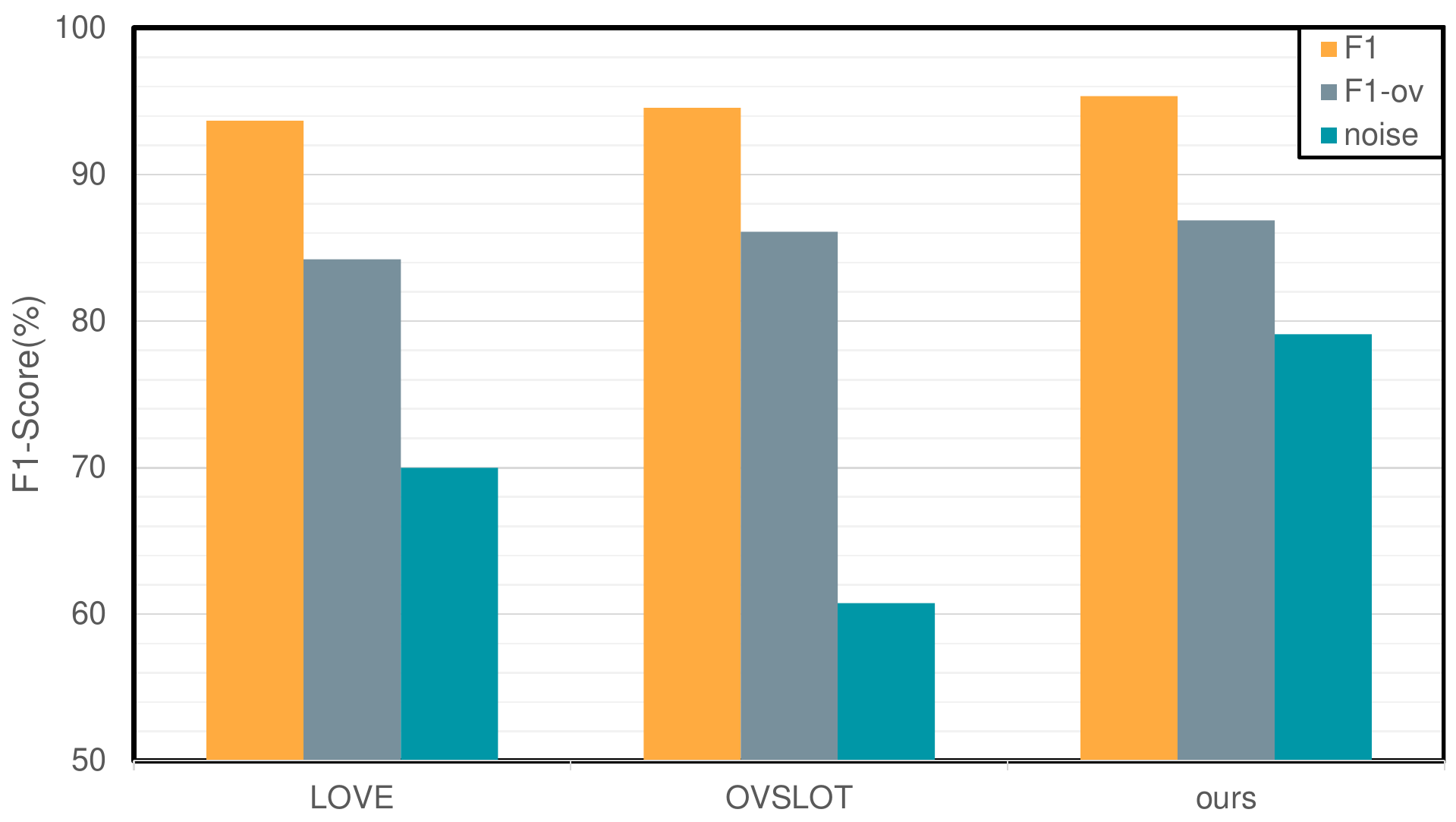}}
\caption{The impact of both OOV slot and OOV word on the slot filling models in real scenarios}
\label{fig:intro2}
\end{figure}


\begin{figure*}[ht]
\centering
\resizebox{0.8\textwidth}{!}{\includegraphics{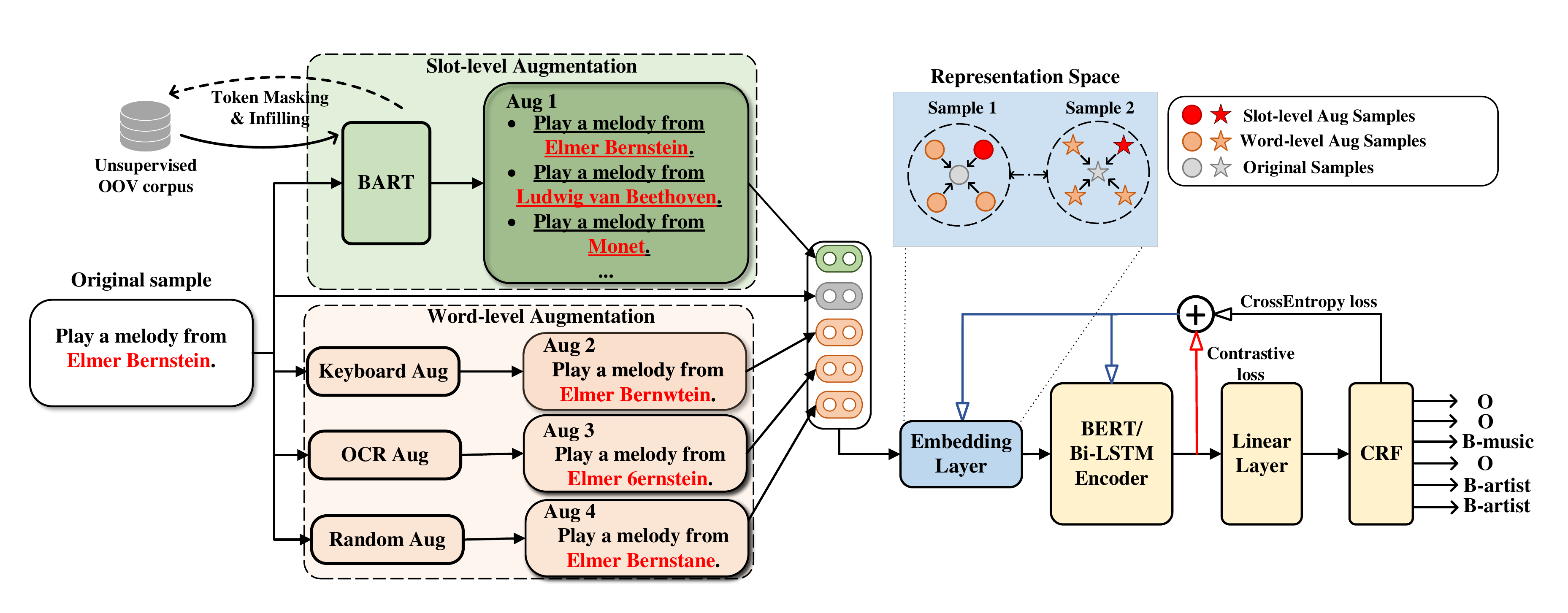}}
\vspace{-0.8cm}
    \caption{The overall architecture of the CMDA framework. Two dotted boxes show the process of two different strategies of data augmentation.}
 
\label{fig:method}
\vspace{-0.3cm}
\end{figure*}


To deal with OOV words problem, FastText\cite{bojanowski2017enriching} pre-trained word embeddings with morphological feature(character n-gram) on large-scale dataset. MIMICK\cite{pinter2017mimicking} and LOVE\cite{chen2022imputing} focused on the surface form of words to generate vectors for OOV words. For OOV slots, Data Noising\cite{kim2019data} add random noise in all slot word embeddings for data augmentation. OVSlotTagging\cite{yan2020adversarial} weakens the role of OOV slot itself through adversarial training. However, all these previous methods only focus on words or slots separately which are limited for practical slot filling task. Figure \ref{fig:intro2} shows that slot filling models performance declines distinctly when facing both word and slot OOV issues. In this paper, we refer to the two problems collectively as OOV problem. As the trend of memorizing entity patterns, these models have poor performance in generalizing to disparate OOV problems. Therefore, it is necessary to design an OOV-robust slot filling model which makes the task-oriented dialogue system more broadly applicable to real scenarios.


To bypass OOV issues, we propose a unified contrastive framework with multi-level data augmentations method named CMDA. We first introduce two strategies of data augmentation which focus on word and slot perspectives, respectively. Specifically, we introduce three word-level augmentation methods, which are Keyboard Augmentation, OCR Augmentation, and Random Augmentation. We intend to guide the model learning OOV word representations from the perspective of word surface form through these three augmentation methods. The slot-level data augmentation guides the model to learn OOV slots distribution by contextual information from unsupervised corpus through a MLM-based training task. Then we propose a multi-contrastive learning framework using these four augmentation data to learn effective representation in feature space. Our main contributions are summarized as follows: (1) To the best of our knowledge, this is the first work to solve OOV problems in both word and slot perspectives. (2) We propose unified contrastive framework with multi-level data augmentations method to deal with the OOV problem by optimize sentence representation. (3) Experimental results demonstrate that our method shows a strong generalization ability in both two OOV problems, and outperforms all previous methods.

\section{Methodology}
\label{sec:method}
We propose two data augmentation strategies to solve the OOV problem, which is shown in detail in Section \ref{sec:aug}. Section \ref{sec:contrastive} provides a comprehensive interpretation about our multi-contrastive learning framework, which uses these augmentation data to learn effective representation by pulling the representations of related sentence pairs together and pushing apart other pairs. 
In Section \ref{sec:train}, we briefly explain the architecture and training objective of our model.

\subsection{Augmentation Strategy}
\label{sec:aug}
\subsubsection{Word-level Augmentation}
In the real world dialogue scene, most OOV words can be linked to in-vocabulary words from the perspective of surface form of word. We aim to pull the hidden representation of OOV word and its corresponding words together so that the model is able to produce high-quality representations which are robust to character perturbations. Thus, we apply character-level perturbations to random tokens in the original sentences to simulate real-world OOV words. We use these new sentences as augmentation samples. To make the distribution of perturbed words more in line with the real-world OOV words, we introduce three novel perturbation methods, resulting in three different word-level augmentation data. \textbf{Keyboard Augmentation}: Simulate keyboard distance error by substituting character; \textbf{OCR Augmentation}: Simulate Optical Character Recognition(OCR) engine error by substituting character; \textbf{Random Augmentation}: Apply augmentation randomly by inserting, substituting, swapping, deleting character. We provide a example of each method in Figure \ref{fig:method}. By this means, we increase the amount and diversity of training samples by using existing data.

\subsubsection{Slot-level Augmentation}

We tend to guide slot filling model to learn the OOV slot semantic structure by absorbing contextual information from an extra open dataset. Inspired by the key idea of the masked language model(MLM\cite{devlin2018bert}), we set a mask-infilling work, which randomly replaces a few tokens in a sentence with the special token [MASK] and recovers the original tokens by a neural network. We use an dataset from the real world with high openness as our unsupervised corpus for training this MLM-based model. Based on this task, we introduce a slot-level augmentation strategy, which is shown in Figure \ref{fig:method}. 

\textbf{Slot-level Augmentation}: We force the generating model to learn OOV slot distribution by randomly masking and infilling tokens of the unsupervised corpus. To transfer the distribution to augmentation data, we mask the slot words of slot filling dataset and apply the model to fill the masked word.
\subsection{Multi-contrastive OOV Objective}
\label{sec:contrastive}


Although we have already transferred the OOV words and slots distribution to the training data by above four types of augmentation data, the representations of these augmentation data are chaotically distributed in the feature space. In order to make better use of these augmentation data, we introduce two contrastive learning framework.
Contrastive learning are used to learn effective representation by pulling positive sentence pairs together and pushing apart negative pairs. 
It assumes a set of paired examples $D=\{s_i, s_i^+\}$, where $s_i$ and $s_i^+$ are original sentence and augmentation sentence. Let $h_i$ and $h_i^+$ denote the representation of $s_i$ and $s_i^+$.

For the word perspective, we employ a supervised contrastive learning(SCL)\cite{khosla2020supervised} objective to learn the robust representation. We formulate SCL as follows:
\begin{equation}
\begin{aligned}
\begin{split}
{\mathcal{L}_{SCL}} = \frac{1}{N}\sum_{i=1}^{N}-\frac{1}{N_{y_i} - 1} \sum_{j=1}^{N_{y_i}}\log \frac{e^{sim(h_i, h_j^+)/\tau_1}}{\sum_{k=1}^{N} \textbf{1}_{i \neq k}e^{sim(h_i, h_k)/\tau_1}}
\end{split}
\end{aligned}
\end{equation}

where $N$ is the total number of examples in the batch and $N_{y_i}$ is the number of positive pairs in the batch. $\tau_1$ is a temperature hyperparameter and $sim(h_1, h_2)$ is cosine similarity $\frac{{h_1^\top} h_2}{||h_1||\cdot||h_2||}$. \textbf{1} is an
indicator function.

For the perspective of OOV slot, we only provide one data augmentation method, which generates one augmentation sample for each sentence. Therefore, we train the contrastive-learning framework by minimizing the InfoNCE\cite{oord2018representation} loss as follow:

\begin{equation}
\begin{aligned}
{\mathcal{L}} = -\frac{1}{N}\sum_{i=1}^{N}\log \frac{e^{sim(h_i, h_i^+)/\tau_2}}{\sum_{j=1}^{N} e^{sim(h_i, h_j)/\tau_2}}
\end{aligned}
\end{equation}
$\tau_2$ is also a temperature hyperparameter. In this way, we take full advantage of the four proposed augmentation methods to transfer the distribution of both OOV words and OOV slots into the sentence feature space. We weight both objectives together as the final objective function for our multi-contrastive learning by a hyperparameter $\alpha$:

\begin{equation}
\begin{aligned}
{\mathcal{L}_{final}} = \alpha \mathcal{L}_{SCL} +(1-\alpha) \mathcal{L}
\end{aligned}
\end{equation}

\subsection{Training and Inference}
\label{sec:train}
The OOV slot filling model consists of a backbone network of Embedding-Encoder-CRF structure. After getting hidden representations of the original data and the four augmented data from the encoder, we bring them into two contrastive learning framework. Then the hidden features are feed into a CRF layer and get the final label of each token. We fine-tune all the parameters using both our multi-contrastive learning objective and cross-entropy loss of the slot filling model.

\begin{table*}[!htbp]\tiny
  \centering
  
   \renewcommand\arraystretch{0.7}
  \resizebox{0.85\textwidth}{!}{
    \begin{tabular}{lcccc|cccc}
    \toprule
    \multicolumn{1}{c}{\multirow{2}[4]{*}{\textbf{\scriptsize Model}}} & \multicolumn{4}{c}{OOV slot}     & \multicolumn{4}{c}{OOV word} \\
\cmidrule{2-9}          & \multicolumn{2}{c}{Snips} & \multicolumn{2}{c}{MR} & \multicolumn{2}{c}{Snips} & \multicolumn{2}{c}{MR} \\
\cmidrule{2-9}          & \multicolumn{1}{c}{F1} & \multicolumn{1}{c}{F1-ov} & \multicolumn{1}{c}{F1} & \multicolumn{1}{c}{F1-ov} & \multicolumn{1}{c}{F1-clean} & \multicolumn{1}{c}{F1-noise} & \multicolumn{1}{c}{F1-clean} & \multicolumn{1}{c}{F1-noise} \\
    \midrule
BiLSTM & 88.99 & 71.78 & 72.07 & 70.39 & 88.99 & 48.91 & 71.78 & 44.22 \\
      {   \space+CRF} & 92.28 & 79.71 & 75.78 & 75.45 & 92.28 & 50.44 & 75.78 & 46.92 \\
     {   \space+random noise} & 92.46 & 82.35 & 75.81 & 75.51 & 92.46 & 50.79 & 75.81 & 47.87 \\
      {   \space+random noise,cw} & 92.89 & 82.58 & 75.92 & 75.60 & 92.89 & 51.76 & 75.92 & 48.20 \\
      { NAT} & 93.01 & - & 76.97 & - & 93.01 & 61.88 & 76.97 & 54.36\\      
      { OVSLOT} & 94.55 & 86.09 & 77.96 & 77.48 & 94.55 & 52.49 & 77.96 & 48.67 \\
      { LOVE} & 93.68 & 83.21 & 76.82 & 75.35 & 93.68 & 59.85 & 76.82 & 54.44 \\
    \textbf{   CMDA} &   \textbf{95.35}   &   \textbf{86.88}    &   \textbf{78.17}    &  \textbf{77.56}     &    \textbf{95.35}   &  \textbf{65.31}    &   \textbf{78.17}   &  \textbf{60.84} \\
    \midrule
    BERT  & 93.31 & 79.77 & 76.07 & 75.40 & 93.31  & 49.62 & 76.07 & 46.73 \\
     {   \space +CRF} & 94.70 & 84.99 & 79.39 & 79.55 & 94.70 & 50.73 & 79.39 & 47.96 \\
      {   \space+random noise} & 95.63 & 87.32 &  79.59 & 79.68 & 95.63 & 51.22 & 79.59 & 48.67 \\
      {   \space+random noise,cw} & 95.57  & 87.18 & 79.49 & 79.56 & 95.57 & 51.98 & 79.49 & 49.39 \\
      {   NAT} & 95.38 & - & 80.86 & - & 95.38 & 63.64 & 80.86 & 57.55 \\
      {   OVSLOT} & 95.87 & 88.06 & 81.61 & 81.78 & 95.87 & 52.88 & 81.61 & 50.44 \\
      {   LOVE} & 95.72 & 87.32 & 81.48 & 79.88 & 95.72 & 61.38 & 81.48 & 56.89 \\
    \textbf{CMDA} &  \textbf{95.96 }    &   \textbf{88.37}    &   \textbf{82.18}    & \textbf{81.92}  &    \textbf{95.96}    &     \textbf{66.37}   &   \textbf{82.18}    & \textbf{63.24} \\
    \bottomrule
    \end{tabular}%
    }
    \caption{The performance on Snips and MR datasets. F1 and F1-clean is the overall score on all slot types, F1-ov is the score on the OOV slots and F1-noise is the score over noise test data. The best results are in bold.}
  \label{tab:experiment}%
\end{table*}%

\section{Experiment}
\label{sec:pagestyle}

\subsection{Datasets}

In the MLM training stage, we employ two multi-modal datasets: Twitter-2015 \cite{zhang2018adaptive}, Twitter-2017 \cite{lu-etal-2018-visual}. We only extract the corpus part and delete the useless details in sentences.
To evaluate CMDA, we conducted experiments on two publicly available benchmark datasets, Snips\cite{coucke2018snips} and MIT-restaurant (MR)\footnote{\url{https://groups.csail.mit.edu/sls/downloads/restaurant/}}. Snips contains user utterances from various domains. MR is a single-domain dataset consisting of restaurant reservations. Both datasets have substantial OOV slots. For the word level, we apply various word-level perturbations from NAT\cite{namysl2020nat} to the test set of both datasets to validate the model performance when facing OOV word problems.

\subsection{Baseline}

We list several baselines for comparison from both word and slot level. For the word-level, we choose Noise-Aware Training (NAT)\cite{namysl2020nat} and LOVE\cite{chen2022imputing} as our baseline. NAT has two novel objectives that can enhance the robustness of sequence labeling, and LOVE which adopts a simple contrastive learning framework makes the word embeddings robust to the OOV word problems. For the slot-level, we compare with CRF, \textit{random noise}, and \textit{random noise, cw}\cite{kim2019data} which means adding random noise in the embeddings of all slot words and concatenating the context word window as input. We also choose OVSLOT\cite{yan2020adversarial} mentioned above as slot-level baseline. We use both the BiLSTM-based\cite{liu2016attention} and BERT-based\cite{devlin2018bert} architecture in slot-filling tasks for a fair comparison.  

\subsection{Implementation Details}
The MLM model of slot-level augmentation is based on BART since the pre-training tasks of BART include token masking and text filling, which are consistent with CMDA. We set the batch size of BART to 8 and set the corresponding learning rate to 1e-5. For our CMDA, we use Glove-BiLSTM as the backbone. The hidden size of BiLSTM is set to 128 and the dropout rate is set to 0.4. We also use BERT to further explore our approach. For all the experiments above, we train and evaluate our model on the NVIDIA RTX 2080Ti GPU.

\subsection{Evaluation}

We evaluate the performance of our approach by using the F1 score\cite{sang2000introduction}. In particular, we adopt F1 scores over perturbed test data generated by the strategy of NAT as F1-noise. For the slot, we follow OVSLOT and adopt the F1 score over all OOV slots, noted as F1-ov, to explore the model's robustness to OOV slot problems. Table \ref{tab:slot} shows all the OOV slots of Snips and MR datasets. We identify the OOV slots based on the diversity of different slot values as well as the average length of slot values.
\begin{table}[htbp]\large
  \centering
  \renewcommand\arraystretch{0.9}
  \resizebox{0.45\textwidth}{!}{
    \begin{tabular}{l|c|c}
    \toprule
    \textbf{Dataset} & \multicolumn{1}{c}{Snips} & \multicolumn{1}{|c}{MR}   \\
    \midrule
    OOV slot & \makecell[c]{playlist, object\_name, entity\_name,\\ album, movie\_name,  track, poi,\\ geographic\_poi, restaurant\_name} &  \makecell[c]{restaurant\_name, dish,\\  amenity, location}      \\
    \bottomrule
    \end{tabular}%
    }
    \caption{The list of all the OOV slots in the two datasets.}
  \label{tab:slot}%
\end{table}%

\vspace{-0.2cm}

\subsection{Main results}
Table \ref{tab:experiment} shows the main results of our method compared to the baselines. Our approach achieves the best performance on both slot-level and word-level issues on two datasets. For the slot-level issue, our proposed method improves by 0.92\% on the F1-ov than OVSLOT on Snips. Meanwhile, for the word-level issue, our approach significantly outperforms the previous state-of-the-art baselines by 9.1\% and 5.5\% on the F1-noise than LOVE and NAT over Snips, 11.8\% and 11.9\% over MR. It demonstrates that benefits from the multi-contrastive learning with joint data augmentation and our approach achieves SOTA in both slot-level and word-level OOV problems. Furthermore, we also experiment with the BERT-based model in our tasks. All the metric scores in BERT-based models outperform the BiLSTM-based models, which indicates that the pretraining models, such as BERT, can effectively capture the context information and can better handle the OOV issues than BiLSTM. 

\vspace{-0.2cm}

\subsection{Ablation Study}

To verify the effectiveness of each part of our framework, we performed experiments on the Snips dataset with our BiLSTM-based model. As shown in Table \ref{tab:ablation},
the absence of slot-level augmentation is more pronounced on F1-ov while F1-noise greatly decrease without word-level augmentation. In this way, we demonstrate that both word-level and slot-level augmentation play a significant role in feature space, missing one part will make the representation lack distribution information, thus make the classification worse.
\begin{table}[htbp]
  \centering
  \small 
  \renewcommand\arraystretch{0.9}
  \resizebox{0.35\textwidth}{!}{
    \begin{tabular}{lccc}
    \toprule
    \textbf{Method} & \multicolumn{1}{l}{\textbf{F1}} & \multicolumn{1}{l}{\textbf{F1-ov}} & \multicolumn{1}{l}{\textbf{F1-noise}}  \\
    \midrule
    CMDA &  95.35  &     86.88  & 65.31 \\
    CMDA w/o word aug & 92.94 & 83.19  &54.73  \\
    CMDA w/o slot aug &  92.74 &  80.88   &  63.68 \\
    \bottomrule
    \end{tabular}%
    }
    \caption{The ablation study results for Snips.}
  \label{tab:ablation}%
\end{table}%

\vspace{-0.4cm}

\section{Conclusion}
\label{sec:print}

We have presented a multi-contrastive learning framework, CMDA, to learn the representations that are robust in the face of out-of-vocabulary problems. We have proposed 4 types of data augmentation methods to improve the robustness of the model from both words and slots perspectives. Through a series of empirical
studies, we have shown that our model can achieve much better performance in predicting both out-of-vocabulary words and slots.

\bibliographystyle{IEEEbib}
\bibliography{strings}

\end{document}